\documentclass[10pt,twocolumn,letterpaper]{article}

\usepackage{iccv}
\usepackage{times}
\usepackage{epsfig}
\usepackage{graphicx}
\usepackage{amsmath,amssymb}
\usepackage{amssymb}
\usepackage{lipsum}
\usepackage{dsfont}
\usepackage{bm}
\usepackage{caption}
\usepackage{subfigure}
\usepackage{multirow}
\usepackage{enumerate}
\usepackage[10pt]{moresize}
\usepackage{xspace}

\usepackage{array}
\newcolumntype{?}{!{\vrule width 2pt}}

\makeatletter
\DeclareRobustCommand\onedot{\futurelet\@let@token\@onedot}
\def\@onedot{\ifx\@let@token.\else.\null\fi\xspace}

\def\eg{{e.g}\onedot} 
\def\ie{{i.e}\onedot}

\makeatother

\newcommand{\METHOD}{GP-GC\xspace} 

\usepackage{url}

\iccvfinalcopy 

\def\iccvPaperID{858} 

\newcommand{\y}{\mathbf{y}}

\newcommand{\M}{\mathbf{M}}
\newcommand{\F}{\mathbf{F}}
\newcommand{\X}{\mathbf{X}}
\newcommand{\I}{\mathbf{I}}
\newcommand{\K}{\mathbf{K}}
\newcommand{\KEps}{\K_\Eps}

\newcommand{\Eps}{\mathcal{E}}
\newcommand{\IEps}{\mathcal{E}^{-1}}

\renewcommand{\i}{\mathrm{i}}
\newcommand{\ii}{\mathrm{ii}}
\newcommand{\iii}{\mathrm{iii}}
\newcommand{\iv}{\mathrm{iv}}

\ificcvfinal\fi
\begin{document}

\title{Identifying Reliable Annotations for Large Scale Image Segmentation}

\author{Alexander Kolesnikov\\
IST Austria\\
{\tt\small akolesnikov@ist.ac.at}
\and
Christoph H. Lampert \\
IST Austria\\
{\tt\small chl@ist.ac.at}
}

\maketitle

\begin{abstract}
Challenging computer vision tasks, in particular semantic image 
segmentation, require large training sets of annotated images. 
While obtaining the actual images is often unproblematic, 
creating the necessary  
annotation is a tedious and costly process. 
Therefore, one often has to work with unreliable annotation sources,  
such as Amazon Mechanical Turk or \mbox{(semi-)}automatic algorithmic 
techniques. 
%

In this work, we present a Gaussian process (GP) based technique for  
simultaneously identifying which images of a training set have 
unreliable annotation and learning a segmentation model in 
which the negative effect of these images is suppressed. 
Alternatively, the model can also just be used to identify the most reliably 
annotated images from the training set, which can then be used for 
training any other segmentation method. 

By relying on "deep features" in combination with a linear covariance
function, our GP can be learned and its hyperparameter determined 
efficiently using only matrix operations and gradient-based optimization. 
This makes our method scalable even to large datasets with several million
training instances. 
\end{abstract}

\section{Introduction}
The recent emergence of large image datasets has led to drastic progress in computer vision. 
In order to achieve state-of-the-art performance for various visual tasks, models are 
trained from millions of annotated images~\cite{krizhevsky2012imagenet,taigman2014deep}.
However, manually creating expert annotation for large datasets requires a tremendous 
amount of resources and is often impractical, even with support by major industrial 
Internet companies.
For example, it has been estimated that creating bounding box annotation for object 
detection tasks takes 25 seconds per box~\cite{su2012crowdsourcing}, and several 
minutes of human effort per image can be required to create pixel-wise annotation 
for semantic image segmentation tasks~\cite{lin2014microsoft}. 

In order to facilitate the data annotation process and make it manageable, 
researchers often utilize sources of annotation that are less reliable but 
that scale more easily to large amounts of data. For example, one harvests 
images from Internet search engines~\cite{schroff2011harvesting} or uses
\emph{Amazon Mechanical Turk (MTurk)} to create annotation. 
Another approach is to create annotation is a (semi-)automatic way, \eg using 
knowledge transfer methods \cite{guillaumin2012large, guillaumin2014imagenet}.

\begin{figure}[t]
        \center
        \includegraphics[width=0.4\textwidth]{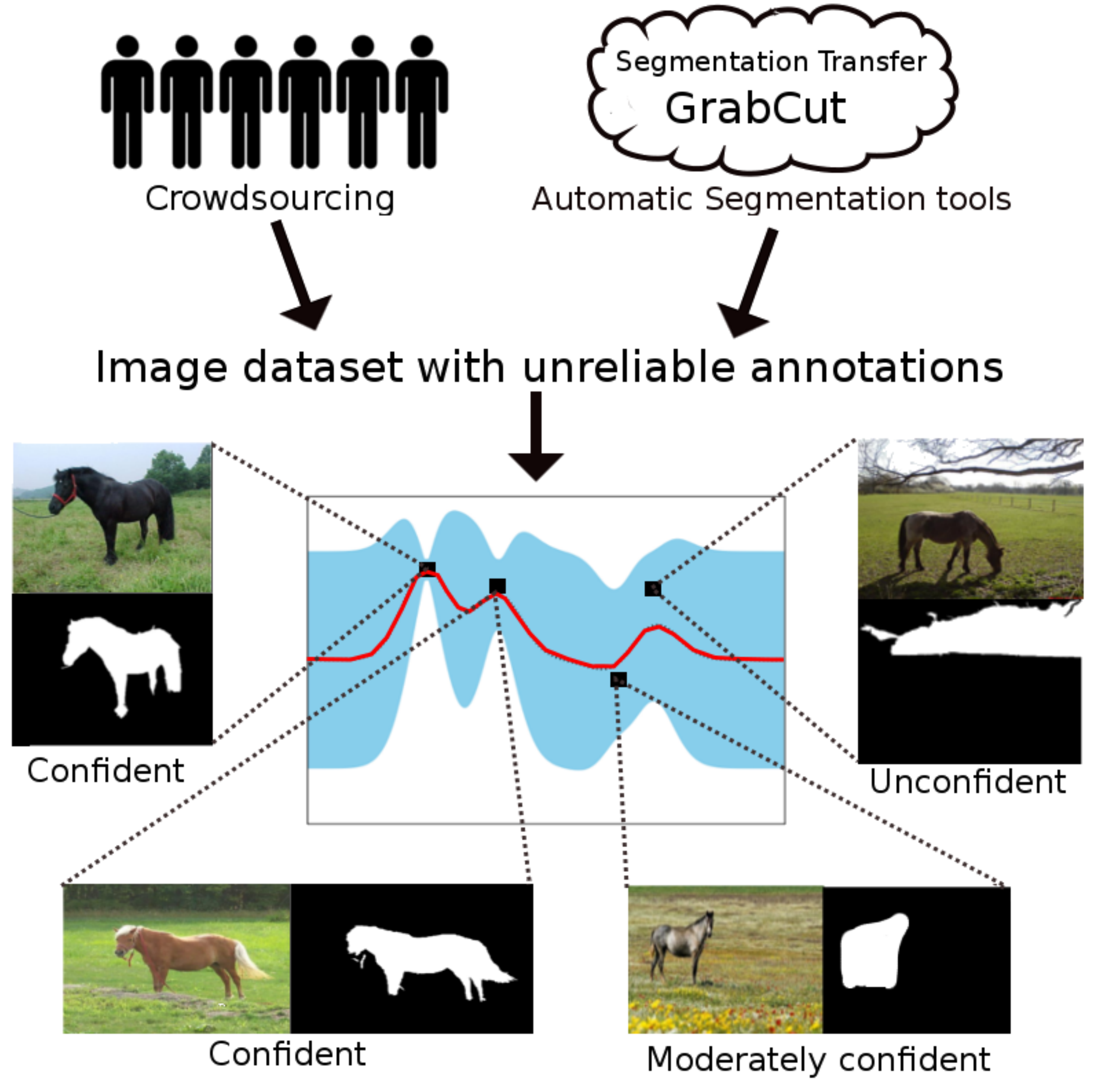}
	\caption{\emph{Learning with unreliable image annotations}: 
             our Gaussian process based method jointly estimates a 
             distribution over prediction models (blue area: 95\% 
             confidence region, 
             red line: most likely model), and a 
             confidence value for the annotation of each image in the 
             training data.}
	\label{fig:teaser}
\end{figure}

A downside of such cheap data sources, in particular automatically 
created annotations, is that they can contain a substantial amount of mistakes. 
Moreover, these mistakes are often strongly correlated: for example, 
MTurk workers will make similar annotation errors in all images 
they handle, 
and an automatic tool, such as segmentation transfer, will work better 
on some classes of images than others. 

Using such noisily annotated data for training can lead to suboptimal 
performance. 
As a consequence, many learning techniques try to identify and suppress 
the wrong or unreliable annotations in the dataset before training. 
However, this leads to a classical chicken-and-egg problem: one needs a 
good data model to identify mislabeled parts of data and one needs 
reliable data to estimate a good model. 

Our contribution is this work is a Gaussian processes (GP)~\cite{rasmussen2006gaussian} 
treatment of the problem of learning with unreliable annotation. 
It avoids the above chicken-and-egg problem by adopting a Bayesian approach,
jointly learning a distribution of suitable models and confidence values for 
each training image (see Figure~\ref{fig:teaser}). 
Afterwards, we use the most likely such model to make predictions. 
All \mbox{(hyper-)}parameters are learned from data, so no model-selection over free parameter,
such as a regularization constant or noise strength, is required.

We also describe an efficient and optionally distributed implementation of 
Gaussian processes with low-rank covariance matrix that scales to segmentation 
datasets with more than 100,000 images (16 million superpixel training instances). 
Conducting experiments on the task of foreground/background image segmentation with large training sets, 
we demonstrate that the proposed method outperforms other approaches for identifying unreliably 
annotated images and that this leads to improved segmentation quality. 

\subsection{Related work}
The problem of unreliable annotation has appeared in the literature 
previously in different contexts. 

For the task of dataset creation, it has become common practice to collect data from unreliable 
sources, such as MTurk, but have each sample annotated by more than one worker 
and combine the obtained labels, \eg by a (weighted) majority vote~\cite{raykar2010learning,sorokin2008utility}.
For segmentation tasks even this strategy can be too costly, and it is 
not clear how annotations could be combined. Instead, it has been suggested to 
have each image annotated only by a single worker, but require workers 
to first fulfill a grading task~\cite{lin2014microsoft}. 
When using images retrieved from search engines, it has been suggested to make 
use of additional available information, \eg keywords, to  filter out mislabeled 
images~\cite{schroff2011harvesting}.

For learning a classifier from unreliable data, the easiest option is to ignore 
the problem and rely on the fact that many discriminative learning techniques 
are to some extent robust against label noise. We use this strategy as one of
the baselines for our experiments in Section~\ref{sec:experiments}, finding 
however that it leads to suboptimal results. 
Alternatively, outlier filtering based on the self-learning heuristic is popular: 
a prediction model is first trained on all data, then its outputs are used to 
identify a subset of the data consistent with the learned model. Afterwards, 
the model is retrained on the subset. 
Optionally, these steps can be repeated multiple times~\cite{deselaers2012weakly}.   
We use this idea as a second baseline for our experiments, showing that it improves 
the performance, but not as much as the method we propose. 

Special variants of popular classification methods, such as support vector machines 
and logistic regression, have been proposed that are more tolerant to label noise 
by explicitly modeling in the objective function the possibility of label changes.
However, these usually result to more difficult optimization problems that need 
to be solved, and they can only be expected to work if certain assumptions about 
the noise are fulfilled, in particular that the label noise is statistically 
independent between different training instances. 
For an in-depth discussion of these and more methods we recommend the recent 
survey on \emph{learning with label noise}~\cite{frenay2014classification}.

Note that recently a method has been proposed by which a classifier is able 
to self-assess the quality of its predictions~\cite{vezhnevets2014associative}. 
While also based on Gaussian processes this work differs significantly 
from ours: it aims at evaluating outputs of a learning system using a 
GP's posterior distribution, while in this work our goal is to assess 
the quality of inputs for a learning system, and we do so using the GP's 
ability to infer hyperparameters from data.

\section{Learning with unreliable annotations}
We are given a training set, $\mathcal{D} = \{(\I_j, \M_j)\}_{j=1}^n$, that consists of 
$n$ pairs of images and segmentations masks.
Each image $\I_j$ is represented as a collection of $r_j$ superpixels, $(x_1, \dots, x_{r_j})$, 
with $x_k \in \mathcal{X}$ for each $k \in \{1 \dots r_j\}$, where $\mathcal{X}$ is 
a universe of superpixels.
Correspondingly, any segmentation mask $\M_j$ is a collection $(y_1, \dots, y_{r_j})$, 
where each $y_j \in \mathcal{Y}$ is the semantic label of the superpixel $x_j$
and $\mathcal{Y}$ is a finite label set. 
For convenience we combine all superpixels and semantic labels from 
the training data and form vectors $\X$ and $\y$ of length $N$,
denoting individual superpixels and semantic labels by a lower index $i$.
In the scope of this work we consider foreground-background segmentation 
problem with $\mathcal{Y} = \{+1,-1\}$, where $+1$ stands for foreground 
and $-1$ for background.
An extension of our technique to the multiclass scenario is possible, but beyond
the scope of this manuscript. 

The main goal of this work is to learn a prediction function, $f: \mathcal{X} \rightarrow \mathcal{Y}$, 
in presence of a significant number of mistakes in the labels of the training data.
We address this learning problem using Gaussian processes.

\subsection{Gaussian processes}

Gaussian processes (GPs) are a prominent Bayesian machine learning technique, 
which in particular is able to reason about noise in data and allows principled, 
gradient-based hyperparameter tuning. 
In this section we reiterate key results from the Gaussian processes literature 
from a practitioner's view. 
For more complete discussion, see~\cite{rasmussen2006gaussian}.

A GP is defined by a positive-definite covariance (or kernel) function, 
$\kappa: \mathcal{X} \times \mathcal{X} \rightarrow \mathbb{R}$, 
that can depend on hyperparameters $\bm{\theta}$. 
For any test input, $\bar{x}$, the GP defines a Gaussian posterior (or predictive) distribution, 
\begin{align}
        p(\bar y|\bar x, \X,\y, \theta) &=  \mathcal{N} \left( m(\bar x), \sigma(\bar x) \right). \label{eq:gp-posterior}
\end{align}
The \emph{mean function},
\begin{align}
        m(\bar x) &= \bar\kappa(\bar x)^\top \K^{-1} \y, \label{eq:gp-posteriormean}
\intertext{allows us to make predictions (by taking its sign), and the \emph{variance}, }
 \sigma(\bar x) &= \kappa(\bar x,\bar x) - \bar\kappa(\bar x)^\top \K^{-1} \bar\kappa(\bar x),         \label{eq:gp-posteriorvariance}
\end{align}
reflects our confidence in this prediction, where $\K$ is the $N\times N$ covariance 
matrix of the training data with entries $\K_{ij}=\kappa(\X_i,\X_j)$ for $i,j\in\{1,\dots,N\}$
and ${\bar{\kappa}(\bar x) = [\kappa(\X_1, \bar{x}), \dots, \kappa(\X_N, \bar{x})]^\top\in\mathbb{R}^{N}}$. 
Note that the mean function \eqref{eq:gp-posteriormean} is the same as one would 
obtain from \emph{kernel ridge regression}~\cite{lampert-fnt2009}, which has 
proven effective also for classification tasks~\cite{rifkin2003regularized}. 

Due to their probabilistic nature, Gaussian processes can incorporate 
uncertainty about labels in the training set. 
One assumes that the label, $y$, of any training example 
is  
perturbed by Gaussian noise with zero mean and variance $\varepsilon^2$. 
Different noise variances for different examples reflect the situation 
in which certain example labels are more trustworthy than others.

The specific form of the GP allows us to integrate out the label 
noise from the posterior distribution. 
The integral can be computed in closed form, resulting in a new 
posterior distribution with mean function,
\begin{align}
        m(\bar x) &= \bar\kappa(\bar x)^\top \KEps^{-1} \y, \label{eq:gp-posteriormean-noisy}
\end{align}
and variance
$\sigma(\bar x) = \kappa(\bar x,\bar x) - \bar\kappa(\bar x)^\top \KEps^{-1} \bar\kappa(\bar x)$, 
for an augmented covariance matrix $\KEps = \K + \mathcal{E}$, where $\mathcal{E}$ 
is the diagonal matrix that contains the noise variances of all training 
examples\footnote{Alternatively, we can think of $\KEps$ as the data 
covariance matrix for a modified covariance function.}.
We consider potential hyperparameters of $\Eps$ as a part of $\bm{\theta}$. 
%

%
%

\subsection{Hyperparameter learning}
A major advantage of GPs over other regression techniques is 
that their probabilistic interpretation offers a principled method for
hyperparameter tuning based on continuous, gradient-based 
optimization instead of partitioning-based techniques such 
as cross-validation. 
%
%
We treat the unknown hyperparameters as random variables and 
study the joint probability $\bm{p}(y,\bm{\theta}|\X)$ over 
hyperparameters and semantic labels. 
Employing type-II likelihood estimation (see~\cite{rasmussen2006gaussian}, chapter 5), 
we obtain optimal hyperparameters, $\bm{\theta}^*$, by solving the following optimization 
problem,
\begin{equation}
        \bm{\theta}^* = \mathrm{argmax}_{\bm{\theta}} \; \ln p(\y | \bm{\theta},\X).
        \label{eq:type-II}
\end{equation}
The expression $\bm{p}(\y|\bm{\theta},\X)$ in the objective~\eqref{eq:type-II} is known as \emph{marginal likelihood}. 
Its value and gradient can be computed in closed form, 
\begin{align}
        &\!\!\ln p(\y | \bm{\theta}, \X) \!=\! -\dfrac{1}{2} \! \left( \y^\top \!\KEps^{-1} \y \!+\! \ln|\KEps| \!+\! N \!\ln(2\pi) \right),
        \label{eq:evidence} \\
        &\dfrac{\partial \ln p(\y | \bm{\theta}, \X)}{\partial \theta} = \dfrac{1}{2} \mathrm{tr} \left( (\alpha \alpha^\top - \KEps^{-1}) \dfrac{\partial \KEps}{\partial \theta} \right),
        \label{eq:evidence-gradient}
\end{align}
where $\alpha=\KEps^{-1} \y$, $\theta$ is any entry of $\bm{\theta}$, 
$\dfrac{\partial \KEps}{\partial \theta}$ is an elementwise partial 
derivative and $|\KEps|$ denotes the determinant of $\KEps$.
If the entries of $\KEps$ depend smoothly on $\bm{\theta}$ then 
the maximization problem \eqref{eq:type-II} is also smooth and 
one can apply standard gradient-based techniques, even for high-dimensional $\bm{\theta}$ (\ie many hyperparameters).
While the solution is not guaranteed to be globally optimal, since~\eqref{eq:type-II} 
is not convex, the procedure has been observed to result in good estimates which 
are largely insensitive to the initialization~\cite{rasmussen2006gaussian}. 



\subsection{A Gaussian process with groupwise confidences}

Our main contribution in this work is a new approach, \METHOD, 
for handling unreliably annotated data in which some training 
examples are more trustworthy than others. 
%
Earlier GP-based approaches either assume that the noise variance 
is constant for all training examples, \ie $\Eps=\lambda\I$ for 
some $\lambda>0$, or that the noise variance is a smooth function 
of the inputs, $\Eps=\text{diag}(g(\X_1),\dots,g(\X_N))$, where $g$ 
is also a Gaussian process function~\cite{goldberg1997regression,kersting2007most}. 
Neither approach is suitable for our situation: constant noise variance 
makes it impossible to distinguish between more and less reliable annotations.
Input-dependent noise variance can reflect only errors due to 
image contents, which is not adequate for errors due to an
unreliable annotation process. 
For example, in image segmentation even identically looking 
superpixels need not share the same noise level if they originate 
from different images or were annotated by different MTurk workers. 

The above insight suggests to allow for arbitrary noise levels, 
$\Eps=\text{diag}(\varepsilon_1,\dots,\varepsilon_N)$, for all 
training instances.
However, without additional constraints this would give too much 
freedom in modelling the data and lead to overfitting. 
Therefore, we propose to take an intermediate route, based on the idea
of estimating label confidence in groups. 
In particular, for image segmentation problems it is sufficient 
to model confidences for the entire image segmentation masks, 
instead of confidences for every individual superpixel.
We obtain such a per-image confidence scores by assuming that 
all superpixel labels from the same image share the same 
confidence value, \ie  $\varepsilon_i = \varepsilon_j$ 
if $\X_i$ and $\X_j$ belong to the same image.
We treat the unknown noise levels as hyperparameters and
learn their value in the way described above. 
Since our confidence about labels is based on the learned 
noise variances, we also refer to the above procedure 
as ``learning label confidence''.
We call the resulting algorithm \emph{Gaussian 
Process with Groupwise Confidences}, or \METHOD.


Note that we avoid the chicken-and-egg problem mentioned in the 
introduction because we simultaneously obtain hyperparameters $\bm{\theta}$,
in particular the noise levels $\bm{\varepsilon} = [\varepsilon_1, \dots, \varepsilon_N]$,
and the predictive distribution.
%
%

\subsection{Instance reweighting}

For unbalanced dataset, \eg in the image segmentation case, where the 
background class is more frequent than the foreground, it makes sense 
to balance the data before training. 
A possible mechanism for this is to duplicate training instances of the 
minority class. Done naively, however, this would unnecessarily increase 
the computational complexity. 
%
Instead, we propose a computational shortcut that allows to incorporate 
duplicate instances without overhead.
Let $w\in\mathbb{N}^N$ be a vector of duplicate counts, \ie $w_i$ is the 
number of copies of the training instance $\X_i$. 
Elementary transformations reveal that for the mean function~\eqref{eq:gp-posteriormean-noisy},
a duplication of training instances is equivalent to changing each hyperparameter
$\varepsilon_i$ to $\varepsilon_i \sqrt{w_i}$. 
We denote vector of hyperparameters, where $\bm{\varepsilon}$ is scaled by squared 
root of entries of $w$ as $\bm{\theta}_w$.
To incorporate duplicates into the marginal likelihood
we also need to scale $\bm{\varepsilon}$ by the square root of 
vector of duplicate counts. In addition, we need to add some terms 
to the marginal likelihood, resulting in the following 
reweighted marginal likelihood,
%
\begin{equation}
	\ln p_w(\y | \bm{\theta}) \hat= \ln p(\y | \bm{\theta}_w) \!+\!
	\dfrac{1}{2}\sum\limits_{i=1}^N [\ln w_i \varepsilon^2_i \!-\! w_i \ln \varepsilon^2_i],
\end{equation}
where ``$\hat=$'' means equality up to a constant that does not depend on the hyperparameters.

Note that the above expressions are well-defined also for non-integer weights, $w$, 
which gives us not only the possibility to increase the importance of samples, 
but also to decrease it, if required. 

\section{Efficient Implementation}\label{sec:linear}

Gaussian processes have a reputation for being computationally demanding. 
Generally, their computational complexity scales cubically with the number 
of training instances and their memory consumption grows quadratically,
because they have to store and invert the augmented data 
covariance matrix, $\KEps$.
Thus, standard implementations of Gaussian processes become 
computationally prohibitive for large-scale datasets. 

Nevertheless, if the sample covariance matrix has a low-rank structure,
all necessary computations can be carried out much faster
by utilizing the Morrison-Sherman-Woodbury identity 
and the matrix determinant lemma~\cite[Corollary 4.3.1]{murphy2012machine}. 
To benefit from this, many techniques for approximating GPs by low-rank GPs 
have been developed using, \eg, the Nystr{\"o}m decomposition~\cite{williams2001using}, 
random subsampling~\cite{drineas2005nystrom}, 
$k$-means clustering~\cite{zhang2008improved}, approximate kernel 
feature maps~\cite{rahimi2007random,vedaldi2012efficient},
or inducing points~\cite{chen2013parallel,quinonero2005unifying}. 

In this work we follow the general trend in computer vision and rely 
on an explicit feature map (obtained from a pre-trained 
\emph{deep network}~\cite{icml2014donahue14,sermanet2013overfeat}) 
in combination with a linear covariance function.
This allows us to develop a parallel and distributed implementation 
of Gaussian Processes with \emph{exact} inference, even in the 
large-scale regime.
Formally, we use a linear covariance function, $\kappa$, of the following form,
\begin{equation}
        \kappa(x_1, x_2) = \phi(x_1)^\top \Sigma\,\phi(x_2),  \label{eq:linear-covariance}
\end{equation}
where $\phi: \mathcal{X} \rightarrow \mathbb{R}^k$ is a $k$-dimensional feature map
with $k\ll N$, and 
$\Sigma = \text{diag}(\sigma_1^2,\dots,\sigma_k^2) \in \mathbb{R}^{k \times k}$ 
is a diagonal matrix of feature scales. 
The entries of $\Sigma$ are assumed to be unknown and included 
in the vector of hyperparameters $\bm{\theta}$.
The feature map $\phi$ induces a feature 
matrix ${\F = [\phi(\X_1), \dots, \phi(\X_N)] \in \mathbb{R}^{k \times N}}$ 
of the training set. 
As a result, the augmented covariance matrix has 
a special structure as sum of a diagonal and a low-rank matrix,
\begin{equation}
\KEps = \mathcal{E} + \F^\top \Sigma \F. \label{eq:KEps-lowrank}
\end{equation}

%
This low-rank representation allows us  
to store $\KEps$ implicitly by storing matrices $\Eps$, $\Sigma$ and $\F$, 
which reduces the memory requirements from $\mathcal{O}(N^2)$ to $\mathcal{O}(Nk)$. 
%
Moreover, all necessary computations for the predictive distribution~\eqref{eq:gp-posterior} 
and marginal likelihood~\eqref{eq:evidence} require only $\mathcal{O}(N k^2)$ 
operations instead of $\mathcal{O}(N^3)$~\cite{williams2001using}. 

Computing the gradients~\eqref{eq:evidence-gradient} with respect to 
unknown hyperparameters generally imposes a computational overhead that 
scales linearly with the number of hyperparameters \cite{quinonero2005unifying,rasmussen2006gaussian}.
For \METHOD, however, we can exploit the homogeneous structure of the hyperparameters, 
$\bm{\theta} = [\bm{\varepsilon}, \bm{\sigma}]$, where $\bm{\varepsilon} = [\varepsilon_1, \dots, \varepsilon_N]$ and 
$\bm{\sigma} = [\sigma_1, \dots, \sigma_k]$ for deriving an expression 
for the gradient without such overhead:
\begin{align}
        &\nabla_{\bm{\varepsilon}} \ln p(\y | \bm{\theta}) =
	\mathrm{diag} \left( (\alpha \alpha^\top - \KEps^{-1}) \mathcal{E}^\prime \right), \\
        &\nabla_{\bm{\sigma}} \ln p(\y | \bm{\theta}) =
	\mathrm{diag} \left( \F (\alpha \alpha^\top - \KEps^{-1}) \F^\top \Sigma^\prime \right),
\end{align}
where $\alpha=\KEps^{-1} \y$ and $\mathcal{E}^\prime$ and $\Sigma^\prime$ are 
diagonal matrices formed by vectors $\bm{\varepsilon}$ and $\bm{\sigma}$ 
respectively.

The computational bottleneck of low-rank Gaussian process learning is 
constituted by standard linear algebra routines, in particular matrix 
multiplication and inversion. 
Thus, a significant reduction in runtime can be achieved by relying on 
multi-threaded linear algebra libraries or even GPUs. 



\begin{figure*}[t]
        \centering
        \subfigure[\emph{HDSeg} dataset: horses (left) and dogs (right).]{\includegraphics[width=\textwidth]{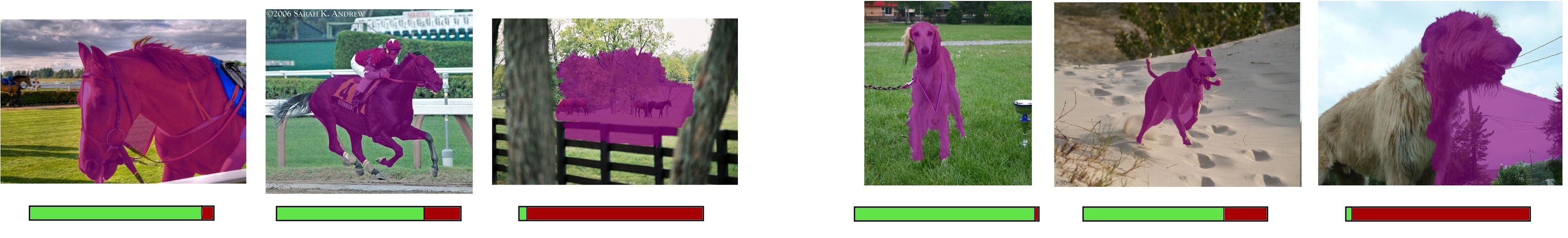}}
	    \subfigure[\emph{AutoSeg} dataset with automatic bounding boxes (depicted in red):
        horses (top left), dogs (top right), cats (bottom left) and sheep (bottom right).]
                  {\includegraphics[width=\textwidth]{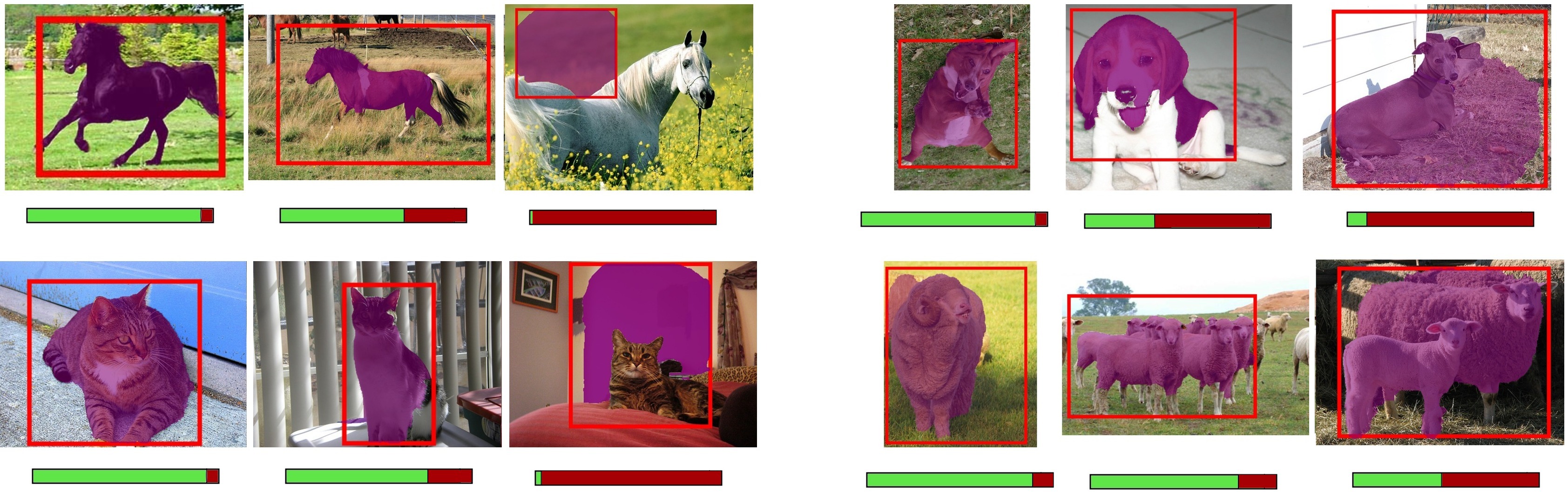}}
        \caption{Examples of training images and their segmentation masks (marked in purple) for the two datasets used. 
                 The horizontal bars reflects the quality value \METHOD assigns to the segmentation masks:
                 the length of the bright green stripe is proportional to the number of images in the 
                 corresponding dataset that are estimated to have lower confidence than the depicted image. }
        \label{fig:datasets}
\end{figure*}

\subsection{Distributed implementation} \label{sec:distributed}

Despite great improvement in performance by utilizing a low-rank covariance function 
and parallel matrix operations, Gaussian processes still remain computationally 
challenging for truly large datasets with high-dimensional feature maps.
For example, one of the datasets we use in our experiments has more than 
100,000 images, 16 million superpixels and 4,113-dimensional feature 
representation. Storing the feature matrix alone requires more than 512~GB RAM,
which is typically not available on a single workstation, but easily achievable 
if the representation is distributed across multiple machines. 

In order to overcome memory limitations and even further improve the 
computational performance we developed a distributed version of 
low-rank Gaussian processes.
It relies on the insight is that the feature matrix $\F$ itself is 
not required for computing the prediction function~\eqref{eq:gp-posteriormean-noisy},
the marginal likelihood \eqref{eq:evidence} and its gradient \eqref{eq:evidence-gradient}, 
if an \emph{oracle} is available for answering the following four queries:
\begin{enumerate}[i.]
        \item compute $\F v$ for any $v \in \mathbb{R}^N$,
        \item compute $\F^\top u$ for any $u \in \mathbb{R}^k$,
        \item compute $\F D \F^\top$ for any diagonal $D \in \mathbb{R}^{N \times N}$,
        \item compute $\mathrm{diag}(\F^\top A \F)$ for any $A \in \mathbb{R}^{k \times k}$.
\end{enumerate}
See Appendix \ref{sec:A} for detailed explanation.
On top of such an oracle we need only $\mathcal{O}(k^2 + N)$ bytes 
of memory and $\mathcal{O}(k^3 +N)$ operations to accomplish all 
necessary computations, which is orders of magnitude faster than 
the original requirements of $\mathcal{O}(Nk)$ bytes 
and $\mathcal{O}(Nk^2)$ operations.

Implementing a distributed version of the oracle is straightforward:
suppose that $p$ computational nodes are available. 
We then split the feature matrix $\F = [\F_1, \F_2, \dots \F_p]$ 
into $p$ roughly equally-sized parts. 
%
Each part is stored on one of the nodes in a distribute way.
All oracle operations naturally decompose with respect to the parts of the feature matrix:
\begin{enumerate}[i.]
	\item ${\F v = \sum_{i=1}^p \F_i v_i}$,
	\item ${\F^\top u = [(\F_1^\top u)^\top, \dots, (\F_p^\top u)^\top]^\top}$,
	\item ${\F D \F^\top = \sum_{i=1}^p \F_i D_i \F_i^\top}$,
    \item ${\mathrm{diag}(\F^\top \!\!A \F) \!=\! [\mathrm{diag}(\F_1^\top \!\!A \F_1)^\top\!\!, \dots,
	                  \mathrm{diag}(\F_p^\top \!A \F_p)^\top]^\top\!\!,}$
\end{enumerate}
where we split the vector $v$ and the diagonal matrix $D$ 
into $p$ parts in the same fashion as we split $\F$, 
obtaining $v_i$ and $D_i$ for all $i \in \{1, \dots, p\}$.
A master node takes care of distributing the objects $v$, $u$, $D$ and $A$ over computational nodes.
Each computational node $i$ calculates $\F_i v_i$, $(\F_i^\top u)^\top$, $\F_i D_i \F_i^\top$
and $\mathrm{diag}(\F_i^\top A \F_i)^\top$
and sends results to the master node,
which collects the results of every operation and aggregates them by taking the 
sum for operations (i) and (iii) or the concatenation for operations (ii) and (iv).
The communication between the master node and computational nodes requires 
sending messages of size at most $\mathcal{O}(k^2 + N)$ bytes, which is 
small in relation to the size of training data.

Consequently, our distributed implementation reduces the time 
and per-machine memory requirements 
by a factor of $p$, at the expense of minor overhead for network communication 
and computations on the master node. 
\section{Experiments}\label{sec:experiments}

\begin{table}[t]
\center
\begin{tabular}{|c|c|c|c|}
	\hline
	Method & SVM  & GP & \METHOD         \\ \hline
    \multicolumn{4}{|c|}{\emph{HDSeg} dataset}     \\ \hline
	Horses (19,060) & 82.5 & 82.5   & \textbf{83.7}       \\ \hline
	Dogs   (111,668) & 80.6 & 80.5   & \textbf{81.3}       \\ \hline
    \multicolumn{4}{|c|}{\emph{AutoSeg} dataset}   \\ \hline
	Horses (9,007) & 81.2 & 80.3   & \textbf{82.5}       \\ \hline
	Dogs  (41,777) & 77.1 & 77.1   & \textbf{79.4}       \\ \hline
	Cats (3,006) & 73.1 & 72.4   & \textbf{73.5}       \\ \hline
	Sheep (5,079) & 75.6 & 75.4   & \textbf{80.0}       \\ \hline
\end{tabular}
\caption{Numerical results (per-class average accuracy in \%) of \METHOD and baseline approaches.
         The numbers in brackets indicate the number of images in the training sets.
         The best numbers are in bold font, see text for details on statistical significance.}
\label{table:segmentation}
\end{table}

We implemented \METHOD in Python, relying the \emph{OpenBlas} 
library\footnote{\scriptsize\url{http://openblas.net}} for linear algebra 
operations and ${\text{L-BFGS}}$~\cite{byrd1995limited} for gradient-based 
optimization. The code will be made publicly available. 

We perform experiments on two large-scale datasets for 
foreground-background image segmentation, see Figure~\ref{fig:datasets} 
for example images. 

1) \emph{HDSeg}~\cite{kolesnikov-eccv2014}\footnote{\scriptsize\url{http://ist.ac.at/~akolesnikov/HDSeg/}}.
We use the 19,060 images of horses and 111,668 images of dogs with segmentation 
masks created automatically by the \emph{segmentation transfer} method~\cite{guillaumin2014imagenet}
for training.
The test images are 241 and 306 manually segmented images of horses and dogs, respectively.

2) \emph{AutoSeg}, a new dataset that we collated from public sources and augmented with 
additional annotations\footnote{We will publish the dataset, including pre-computed features.}.
%
%
The training images for this dataset are taken from the \emph{ImageNet} project\footnote{\scriptsize\url{http://www.image-net.org}}. 
There are four categories: horses (9,007 images), dogs (41,777 images), cats (3,006 images), sheep (5,079 images).
All training images are annotated with segmentation masks generated automatically by the \emph{GrabCut} algorithm~\cite{rother2004grabcut}
%
from the \emph{OpenCV} library\footnote{\scriptsize\url{http://opencv.org}} with default parameters.
We initialize \emph{GrabCut} with bounding boxes that were also generated automatically by the 
\emph{ImageNet Auto-Annotation} method~\cite{vezhnevets2014associative}\footnote{\scriptsize\url{http://groups.inf.ed.ac.uk/calvin/proj-imagenet/page}}.
%
The test set consist of 1001 images of horses, 1521 images of dogs, 1480 images of cats and 489 images of sheep
with manually created per-pixel segmentation masks that were taken from the validation part of 
the \emph{MS\,COCO}~\footnote{\scriptsize\url{http://mscoco.org/}} dataset.
%

As evaluation metric for both datasets we use the average class accuracy~\cite{kolesnikov-eccv2014}:
we compute the percentage of correctly classified foreground pixels and the 
percentage of correctly classified background pixels across all images and average both values. 
To assess the significance of reported results, the above single number is not sufficient.
Therefore, we use a closely related quantity for this purpose: we compute an average class 
accuracy as above separately for every image and perform a Wilcoxon signed-rank test \cite{wilcoxon1945individual} with significance 
level $10^{-3}$. 



\setlength\tabcolsep{2pt}
\begin{table}[t]
\center
\begin{tabular}{|c|}
               \hline
               Prediction \\ 
               model \\ \hline 
               Selection \\ 
               rule  \\ \hline 
    	       \\ \hline 
	       Horses \\ 
	       Dogs   \\ \hline 
    	      \\ \hline 
	       Horses \\ 
	       Dogs   \\ 
	       Cats   \\ 
           Sheep  \\ \hline 
\end{tabular}
\;
\begin{tabular}{|c|c|}
               \hline
               \multicolumn{2}{|c|}{SVM} \\
               \multicolumn{2}{|c|}{Classifier} \\ \hline
               SVM  & \METHOD \\
               margin  & confidence \\ \hline
    	       \multicolumn{2}{|c|}{\emph{HDSeg} dataset}     \\ \hline
	           83.8  &  \textbf{84.3}   \\
	           81.7  &  \textbf{82.0}   \\ \hline
    	       \multicolumn{2}{|c|}{\emph{AutoSeg} dataset}     \\ \hline
	           82.5  &  \textbf{83.2}   \\
	           79.2  &  \textbf{80.9}   \\
	           71.9  &  \textbf{72.9}   \\
               80.2  &  \textbf{81.7}   \\ \hline       
\end{tabular}
\;
\begin{tabular}{|c|c|}
               \hline
               \multicolumn{2}{|c|}{GP} \\
               \multicolumn{2}{|c|}{Classifier} \\ \hline
                  SVM  & \METHOD \\
                  margin  & confidence \\ \hline
    	       \multicolumn{2}{|c|}{\emph{HDSeg} dataset}     \\ \hline
            83.5  &  \textbf{84.3}   \\
	           81.2  &  \textbf{81.7}   \\ \hline
    	       \multicolumn{2}{|c|}{\emph{AutoSeg} dataset}     \\ \hline
	         82.7  &  \textbf{83.9}   \\
	          79.7  &  \textbf{81.2}   \\
	          72.5  &  \textbf{73.9}   \\
	          81.2  &  \textbf{82.9}   \\ \hline       
\end{tabular}
\caption{Numerical results (per-class average accuracy in \%) of training an SVM or GP on filtered data.
         An SVM or GP classifier is trained on 25\% of the most reliable images from the dataset, as 
         selected by the SVM margin or \METHOD filtering.
         The best numbers are in bold font, see text for details on statistical significance.}
\label{table:comp-baseline}
\end{table}

\setlength\tabcolsep{5pt}
\begin{table*}[t]
\centering
\begin{tabular}{|c|c?c|c|c|c|c|c|c|c|}
	\hline
    Method & \METHOD          & Top-1\%  & Top-2\%  & Top-5\%  & Top-10\%  & Top-15\%  & Top-25\%  & Top-50\% & Top-75\% \\ \hline
    \multicolumn{10}{|c|}{\emph{HDSeg} dataset} \\ \hline
    Horses & 83.7     & 82.7     & 83.0     & 83.7     & 83.9      & 84.1    & 84.3    & \textbf{84.4}   & 83.9   \\ \hline
    Dogs   & 81.3     & 80.0     & 80.3     & 80.8     & 81.2      & 81.4    & 81.7    & \textbf{81.9}   & 81.6   \\ \hline
    \multicolumn{10}{|c|}{\  \emph{AutoSeg} dataset} \\ \hline
    Horses & 82.5     & 82.4     & 82.8     & 83.3     & 83.8      & 83.8    & \textbf{83.9}    & 83.4   & 83.6   \\ \hline
    Dogs   & 79.4     & 77.1     & 77.7     & 78.9     & 80.1      & 80.7    & \textbf{81.2}    & 80.8   & 79.9   \\ \hline
    Cats   & 73.5     & 57.0     & 69.3     & 72.3     & 72.6      & 73.3    & 73.9    & \textbf{74.4}   & 74.3   \\ \hline
    Sheep  & 80.0     & 78.1     & 80.4     & 81.6     & 82.6      & 82.5    & \textbf{82.9}    & 81.4   & 79.4   \\ \hline
\end{tabular}
\caption{Numerical results (per-class average accuracy in \%) of training \emph{GP} model from different subsets of training data.
         Column ``Top-$\gamma$\%'' indicates that we select $\gamma$\% of the most reliable images
         according to confidences learned by \METHOD. The best numbers are in bold font.}
\label{table:filtering}
\end{table*}

\subsection{Image Representation}\label{sec:features}
We split every image into superpixels using the \emph{SLIC} \cite{achanta2012slic} method from \emph{scikit-image}\footnote{\scriptsize\url{http://scikit-image.org}} library.
Each superpixel is assigned a semantic label based on the majority vote of pixel labels inside it. 
For each superpixel we compute appearance-based features using the \emph{OverFeat}\cite{sermanet2013overfeat}
library\footnote{\scriptsize\url{http://cilvr.nyu.edu/doku.php?id=software:overfeat:start}}. 
We extract a 4096-dimensional vector from the output of the 20th layer of the pretrained 
model referred to as \emph{fast} model in the library documentation.
%
%
%
Additionally, we add features that describe the position of a superpixel in its image.
For this we split each image into a 4x4 uniform grid and describe position of each superpixels by 16 values,
 each one specifying the ratio of pixels from the superpixel falling into the corresponding grid cell.
We also add a constant (bias) feature, 
resulting in an overall feature map, $\phi: \mathcal{X} \rightarrow \mathbb{R}^{k}$, with $k=4113$. 
%
%
The features within each of the three homogeneous groups (appearance, position, constant) share 
the same scale hyperparameter in the covariance function~\eqref{eq:linear-covariance},
\ie $\sigma_i=\sigma_j$ if the feature dimensions $i$ and $j$ are within the same group. 

\subsection{Baseline approaches}\label{sec:baseline}

We compare \METHOD against two baselines.
The first baseline is also a Gaussian process, but we 
assume that all superpixels have the same noise variance. 
All hyperparameters are again learned by type-II maximum likelihood. 
We refer this method simply as GP. 
This baseline is meant to study if a selective estimation of the confidence values indeed has a positive effect on prediction performance.

As second baseline we use a linear support vector machine (SVM), relying 
on the \emph{LibLinear} implementation with squared slack variables, which 
is known to deliver state-of-the-art optimization speed and prediction quality~\cite{fan2008liblinear}.
For training SVM models we always perform 5-fold cross-validation to determine the regularization constant $C\in \{2^{-20}, 2^{-19},\dots, 2^{-1}\}$.

\subsection{Foreground-background Segmentation}

We conduct experiments on the \emph{HDSeg} and \emph{AutoSeg} datasets, 
analyzing the potential of \METHOD for two tasks: either as a dedicated method 
for semantic image segmentation, or as a tool for identifying reliably 
annotated images, which can be used afterwards, \eg, as a training set 
for other approaches. 
For all experiments we reweight training data 
so that foreground and background classes are balanced and all instances with 
the same semantic label have the same weight, but the overall weight remains 
unchanged, \ie $\sum_i w_i=N$. 
This step removes the effect of different ratios of foreground and background labels 
for different datasets and their subsets. 

The first set of experiments compares \METHOD with the baselines, GP and SVM, 
on the task of foreground-background segmentation. 
Numeric results are presented in Table~\ref{table:segmentation}. 
They show that \METHOD achieves best results for all datasets and all semantic classes. 
According to a Wilcoxon signed-rank test, \METHOD's improvement over the baselines 
is significant to the $10^{-3}$ level in all cases. 

We obtain two insights from this. First, the fact that \METHOD improves over GP confirms 
that it is indeed beneficial to learn different confidence hyperparameters for different images.
Second, the results also confirm 
that classification using Gaussian 
process regression with gradient-based hyperparameter selection yields results comparable with 
other state-of-the-art classifiers, such as SVMs, whose regularization parameter have to be 
chosen by more tedious cross-validation procedures.  


In a second set of experiments we benchmark \METHOD's ability to suppress images with unreliable 
annotation. 
For this, we apply \METHOD to the complete training set and use the learned hyperparameter 
values (see Figure~\ref{fig:datasets} for an illustration) to form a new data set that 
consists only of the 25\% of images that \METHOD was most confident about. 
We compare this approach to SVM-based filtering similar to what has 
been done in the computer vision literature before~\cite{deselaers2012weakly}:
we train an SVM on the original dataset and form per-image confidence values 
by averaging the SVM margins of the contained superpixels. Afterwards we use 
the same construction as above, forming a new data set from the 25\% of images 
with highest confidence scores. 

We benchmark how useful the resulting data sets are by using them as training sets for 
either a GP (with single noise variance) or an SVM. Table~\ref{table:comp-baseline} shows 
the results. 
By comparing the results to Table~\ref{table:segmentation}, one sees that both methods for 
filtering out images with unreliable annotation help the segmentation accuracy. 
However, the improvement from filtering using \METHOD is higher than when using the 
data filtered by the SVM approach, regardless of the classifiers used afterwards.
This indicates that \METHOD is a more reliable method for suppressing bad annotation. 
According to a Wilcoxon test, \METHOD's improvement over the other method is significant to 
the $10^{-3}$ level in 11 out of 12 cases (all except \emph{AutoSeg sheep} for the SVM classifier).


To understand this effect in more detail, we performed another experiment: we used \METHOD 
to create training sets of different sizes (1\% to 75\% of the original training sets) and 
trained the GP model on each of them. 
The results in Table~\ref{table:filtering} show that the best results are consistently 
obtained when using 25\%--50\% of the data.
For example, for the largest dataset (\emph{HDSeg~dog}), the quality of the 
prediction model keeps increasing up to a training set of over 55,000 
images (8 million superpixels). 
This shows that having many training images (even with unreliable annotations) is beneficial 
for the overall performance and that scalability is an important feature of our approach. 

\section{Summary}\label{sec:discussion}

In this work we presented, \METHOD, an efficient and parameter-free method 
for learning from datasets with unreliable annotation, in particular 
for image segmentation tasks. 
The main idea is to use a Gaussian process to jointly model the 
prediction model and confidence scores of individual annotation in 
the training data.
The confidence values are shared within groups of examples, \eg all 
superpixels within an image, and can be obtained automatically using 
Bayesian reasoning and gradient-based hyperparameter tuning. 
As a consequence there are no free parameter that need to be tuned. 

In experiments on two large-scale image segmentation 
datasets, we showed that by learning individual confidence values \METHOD 
is able to better cope with unreliable annotation than other 
classification methods.
Furthermore, we showed that the estimated confidences allow us to filter
out examples with unreliable annotation, thereby providing a way to 
create a cleaner dataset that can afterwards be used also by 
other learning methods. 

By relying on an explicit feature map and low-rank kernel, \METHOD training 
is very efficient and easily implemented in a parallel or even distributed 
way. For example, training with 20 machines on the \emph{HDSeg~dog} segmentation dataset, 
which consists of over 100,000 images (16 million superpixels), takes only 
a few hours. 

%
%
%
%

\appendix
\section{Reduction to the oracle}\label{sec:A}

We demonstrate that having the \emph{oracle} from Section \ref{sec:distributed}
is sufficient to compute the mean function \eqref{eq:gp-posteriormean}, 
the marginal likelihood \eqref{eq:evidence}, and its 
gradient \eqref{eq:evidence-gradient} without access to the 
feature matrix $\F$ itself.
We highlight terms that \emph{oracle} can compute by braces with the number 
of the corresponding \emph{oracle} operation.

We first apply the Sherman-Morrison-Woodbury identity and matrix determinant lemma to the matrix $\KEps$:  
\begin{align}
	&{\KEps^{-1} \!=\! (\Eps + \F^\top \Sigma \F)^{-1} \!=\! \IEps \!-\! \IEps \F^\top C^{-1} \F \IEps}, \label{eq:Woodbury} \\
        &\!\!\ln\! |\KEps| \!=\! \ln\! |\Eps \!+\! \F^\top \Sigma \F| \!=\! \ln\!|\Eps| \!+\! \ln\!|\Sigma| \!+\! \ln\!|C|, \label{eq:determinant}
\end{align}
where we denote $C = \Sigma^{-1} + \underbrace{\F \IEps \F^\top}_{(iii)}$.

For convenience we introduce $\tilde{y} = \IEps y$.
Relying on \eqref{eq:Woodbury} we compute the following expressions:
\begin{align}
	&y^\top \KEps^{-1} y = y^\top \tilde{y} -
	{\underbrace{(\F \tilde{y})}_{(\i)}}^\top  C^{-1} \underbrace{(\F \tilde{y})}_{(\i)}, \label{eq:yKy}\\
        &\F \KEps^{-1} y = \underbrace{\F \tilde{y}}_{(\i)} - \underbrace{(\F \IEps \F^\top)}_{(\iii)}
	C^{-1} \underbrace{(\F \tilde{y})}_{(\i)}, \label{eq:FKy} \\
	&\F \KEps^{-1} \F^\top = \underbrace{(\F \IEps \F^\top)}_{(\iii)}
	                      (\I - C^{-1} \underbrace{(\F \IEps \F^\top)}_{(\iii)}), \label{eq:FKF} \\
	&\alpha = \KEps^{-1} y = \tilde{y} -
	\IEps \overbrace{\F^\top C^{-1} \underbrace{\F \tilde{y}}_{(\i)}}^{(\ii)}, \\
	&\mathrm{diag}(\KEps^{-1}) = \mathrm{diag}(\Eps^{-1}) - \label{eq:K_diag}\\
	&\;\;\;\;\;\;\;\;\;\;\;\;\;\;\;\;\;\;\;\;\;
	 \underbrace{\mathrm{diag}(\F^\top C^{-1} \F)}_{(\iv)} \odot \mathrm{diag}(\Eps^{-2}), \notag
 \end{align}
where $\odot$ is elementwise (Hadamard) vector multiplication.

Using the above identities, we obtain the mean of the predictive distribution,
\begin{equation}
	m(\bar{x}) = \bar{\kappa}(\bar x)^\top \KEps^{-1} y = \phi(\bar{x})^\top \underbrace{\F \KEps^{-1} y}_{\eqref{eq:FKy}},
\end{equation}
and the marginal likelihood, 
\begin{equation}
	\ln p(\y | \bm{\theta}) \!=\! -\dfrac{1}{2} ( \underbrace{\y^\top \KEps^{-1} \y}_{\eqref{eq:yKy}} + 
	\underbrace{\ln|\KEps|}_{\eqref{eq:determinant}} + N\! \ln(2\pi) ).
\end{equation}

Finally, we compute the gradient of the marginal likelihood with respect 
to the noise variances $\bm{\varepsilon}$,
\begin{align}
	&\nabla_{\bm{\varepsilon}} \ln p(\y | \bm{\theta}) = \mathrm{diag} \left( (\alpha \alpha^\top - \KEps^{-1}) \Eps^\prime \right) \\
        & = (\alpha \odot \alpha - \underbrace{\mathrm{diag}(\KEps^{-1})}_{\eqref{eq:K_diag}}) \odot \mathrm{diag}(\Eps^\prime) \notag
\end{align}
and with respect to the feature scales $\bm{\sigma}$,
\begin{align}
	&\nabla_{\bm{\sigma}} \ln p(\y | \bm{\theta}) = \mathrm{diag} \left( \F (\alpha \alpha^\top - \KEps^{-1}) \F^\top \Sigma^\prime \right) \\
	       & = (\underbrace{(\F \alpha)}_{(\i)} \odot \underbrace{(\F \alpha)}_{(\i)} - \mathrm{diag}(\underbrace{\F \KEps^{-1} \F^\top}_{\eqref{eq:FKF}})) \odot \mathrm{diag}(\Sigma^\prime). \notag
\end{align}

{\small
\bibliographystyle{ieee}
\bibliography{paper}
}

\end{document}